\documentclass[10pt,twocolumn,letterpaper]{article}
\pdfoutput=1

\usepackage[pagenumbers]{cvpr} 

\usepackage{graphicx}
\usepackage{amsmath}
\usepackage{amssymb}
\usepackage{booktabs}
\usepackage{adjustbox}
\usepackage{comment}

\usepackage[pagebackref,breaklinks,colorlinks]{hyperref}

\usepackage[capitalize]{cleveref}
\crefname{section}{Sec.}{Secs.}
\Crefname{section}{Section}{Sections}
\Crefname{table}{Table}{Tables}
\crefname{table}{Tab.}{Tabs.}

\newcommand\blfootnote[1]{%
  \begingroup
  \renewcommand\thefootnote{}\footnote{#1}%
  \addtocounter{footnote}{-1}%
  \endgroup
}

\begin{document}

\title{NICE: CVPR 2023 Challenge on Zero-shot Image Captioning}
\author{Taehoon Kim\\
\and
Pyunghwan Ahn\\
\and
Sangyun Kim\\
\and
Sihaeng Lee\\
\and
Mark Marsden\\
\and
Alessandra Sala\\
\and
Seung Hwan Kim\\
\and
Bohyung Han\\
\and
Kyoung Mu Lee\\
\and
Honglak Lee\\
\and
Kyounghoon Bae\\
\and
Xiangyu Wu\\
\and
Yi Gao\\
\and
Hailiang Zhang\\
\and
Yang Yang\\
\and
Weili Guo\\
\and
Jianfeng Lu\\
\and
Youngtaek Oh\\
\and
Jae Won Cho\\
\and
Dong-Jin Kim\\
\and
In So Kweon\\
\and
Junmo Kim\\
\and
Wooyoung Kang\\
\and
Won Young Jhoo\\
\and
Byungseok Roh\\
\and
Jonghwan Mun\\
\and
Solgil Oh\\
\and
Kenan Emir Ak\\
\and
Gwang-Gook Lee\\
\and
Yan Xu\\
\and
Mingwei Shen\\
\and
Kyomin Hwang\\
\and
Wonsik Shin\\
\and
Kamin Lee\\
\and
Wonhark Park\\
\and
Dongkwan Lee\\
\and
Nojun Kwak\\
\and
Yujin Wang\\
\and
Yimu Wang\\
\and
Tiancheng Gu\\
\and
Xingchang Lv\\
\and
Mingmao Sun\\
}
\maketitle
\blfootnote{Affiliation and role of each author are listed in the appendix.}

\begin{figure*}[ht]
    \centering
    \begin{subfigure}[b]{0.24\textwidth}
        \centering
        \includegraphics[width=\textwidth]{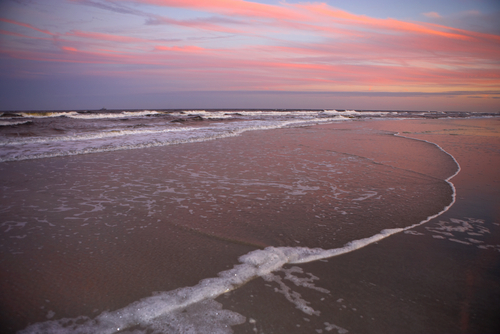}
        \caption{View from sandy beach of picturesque sunset horizon over sea surf lapping on shore}
        \label{fig:nice1}
    \end{subfigure}
    \hfill
    \begin{subfigure}[b]{0.24\textwidth}
        \centering
        \includegraphics[width=\textwidth]{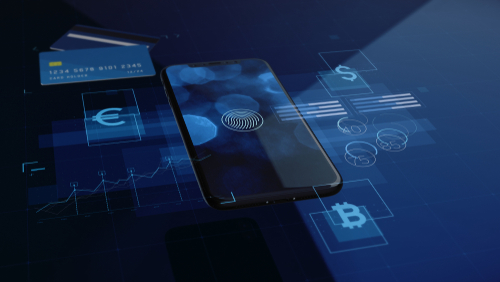}
        \caption{3D concept of online banking on mobile phone showing services currencies and graphs}
        \label{fig:nice2}
    \end{subfigure}
    \hfill
    \begin{subfigure}[b]{0.2\textwidth}
        \centering
        \includegraphics[width=\textwidth]{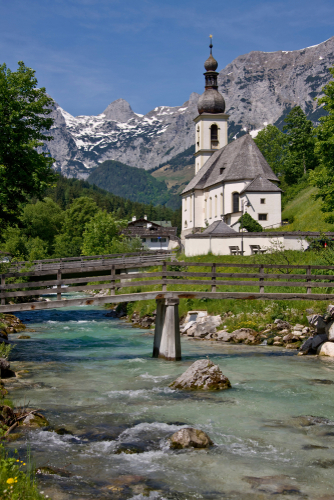}
        \caption{St. Sebastian’s Church, Ramsau, near Berchtesgaden, Bavaria, Germany}
        \label{fig:nice3}
    \end{subfigure}
    \hfill
    \begin{subfigure}[b]{0.24\textwidth}
        \centering
        \includegraphics[width=\textwidth]{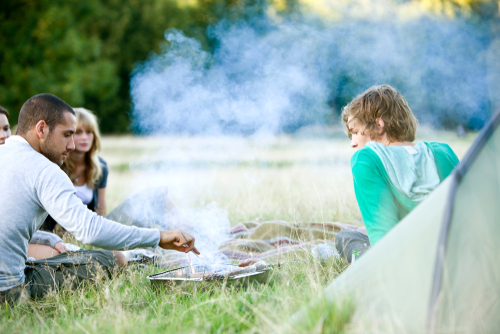}
        \caption{A group of young friends having a barbecue outdoors next to a tent}
        \label{fig:nice4}
    \end{subfigure}
    \hfill
    \caption{Example images and corresponding captions in NICE dataset.}
    \label{fig:nice}
\end{figure*}

\begin{abstract}
   In this report, we introduce NICE (New frontiers for zero-shot Image Captioning Evaluation) project\footnote{\url{https://nice.lgresearch.ai/}} and share the results and outcomes of 2023 challenge. This project is designed to challenge the computer vision community to develop robust image captioning models that advance the state-of-the-art both in terms of accuracy and fairness. Through the challenge, the image captioning models were tested using a new evaluation dataset that includes a large variety of visual concepts from many domains. There was no specific training data provided for the challenge, and therefore the challenge entries were required to adapt to new types of image descriptions that had not been seen during training. This report includes information on the newly proposed NICE dataset, evaluation methods, challenge results, and technical details of top-ranking entries. We expect that the outcomes of the challenge will contribute to the improvement of AI models on various vision-language tasks.
\end{abstract}

\section{Introduction}\label{sec:intro}

Zero-shot image captioning is a task that general purpose vision-language models must perform well, as it requires both the visual understanding of the scene and the ability to describe it in natural language. Automatic generation of image descriptions makes a variety of applications possible, such as improved image search with the help of natural language, better detection of inappropriate contents on the web, and explanation of visual contents to visually impaired people. In most real-world scenarios, images from unseen environments are frequently given to perform these actions, which makes zero-shot image captioning essential.

In the earlier days, image captioning models were trained on curated datasets which consist of training and testing data from the same domain or categories~\cite{chen2015microsoft, sidorov2020textcaps}. These models inevitably have limited capability to recognize a whole new concept and describe it. For better utility of image captioning models, several datasets~\cite{agrawal2019nocaps} were suggested in order to test the model with images from unseen categories. This evaluation process has put the existing models on harder testing conditions, leading to development of image captioning models that can understand general scenes.

Although there are a few benchmarks which target for zero-shot image captioning evaluation, each of them lacks in one or more requirements, which includes large size of the dataset, variety of categories, and quality of language descriptions. The size of the dataset is essential to test the models with adequate number of images to guarantee reliability, and variety of categories is required so that the tested models are not fitted to perform well on a few certain concepts. Also, as the prediction shall be usually compared with the ground truth captions using text comparison metrics, the quality of the ground truth captions must be guaranteed to be accurate and of high quality.

Through NICE 2023 challenge, we made a newly curated NICE dataset, which consists of 26k images and corresponding high-quality captions, publicly available. Furthermore, we did not provide a specific training data for the challenge, which forces the models to be trained to generalize well, thus achieving zero-shot capability. Even though this challenge was a newly organized one, 51 and 31 teams competed in the validation and test phase, respectively, and the top-scoring entries showed very small differences in the final score. The following sections of the report includes the information on the challenge, evaluation methods, results, and proposed approaches by top-scoring entries.


\section{Challenge}\label{sec:chal}

In this challenge, a newly curated dataset was made publicly available, named NICE dataset, and then the challenge was organized to evaluate image captioning capability of AI models. This section introduces the dataset, evaluation method, challenge phases, and results.

\begin{table*}[ht]
\centering
\begin{adjustbox}{width=\textwidth}
\begin{tabular}{|l|l|l|l|l|l|l|l|l|l|}
\hline
Rank & Username     & Bleu\_1      & Bleu\_2      & Bleu\_3      & Bleu\_4      & ROUGE\_L     & CIDEr         & METEOR       & SPICE        \\ \hline
1 & jinx         & 56.0839 (3)  & 46.5881 (2)  & 40.0586 (2)  & 35.1730 (2)  & 55.5685 (2)  & 325.7216 (1)  & 29.1455 (2)  & 44.4351 (2)  \\ \hline
2 & stack-top    & 58.0129 (1)  & 47.8769 (1)  & 40.9018 (1)  & 35.7796 (1)  & 56.3780 (1)  & 324.9277 (2)  & 30.0329 (1)  & 45.5456 (1)  \\ \hline
3 & PEJI         & 56.4908 (2)  & 46.5371 (3)  & 39.6859 (3)  & 34.5996 (3)  & 54.9832 (3)  & 316.2290 (3)  & 28.9407 (3)  & 43.8281 (3)  \\ \hline
4 & calisolo     & 55.7849 (4)  & 45.4753 (4)  & 38.4468 (4)  & 33.1970 (4)  & 52.9229 (4)  & 287.6926 (4)  & 27.9402 (4)  & 41.3584 (4)  \\ \hline
5 & img\_capt    & 53.6418 (7)  & 42.8248 (7)  & 35.6963 (6)  & 30.5766 (5)  & 51.6134 (7)  & 278.1607 (5)  & 27.0087 (7)  & 40.9747 (6)  \\ \hline
6 & kyominhwang  & 54.1975 (6)  & 43.0211 (6)  & 35.4753 (7)  & 30.0185 (7)  & 52.3251 (5)  & 274.6941 (6)  & 27.3464 (5)  & 41.0430 (5)  \\ \hline
7 & Mtop         & 54.4547 (5)  & 43.3951 (5)  & 35.8492 (5)  & 30.2877 (6)  & 52.1572 (6)  & 270.5980 (7)  & 27.2156 (6)  & 40.7732 (7)  \\ \hline
8 & Yongsik      & 51.3670 (8)  & 40.2552 (8)  & 32.7835 (8)  & 27.4779 (8)  & 50.5170 (8)  & 255.9013 (8)  & 26.1637 (8)  & 39.4600 (8)  \\ \hline
9 & TXHmercury   & 50.7199 (10) & 39.0901 (11) & 31.1348 (12) & 25.3470 (13) & 50.0612 (9)  & 239.0106 (9)  & 25.3394 (9)  & 38.5370 (9)  \\ \hline
10 & zero\_score  & 50.1327 (12) & 39.4831 (9)  & 32.4496 (9)  & 27.4089 (9)  & 48.0671 (12) & 238.3091 (10) & 24.8445 (12) & 37.0757 (10) \\ \hline
11 & sungbin.son  & 48.2634 (14) & 37.8367 (13) & 31.0821 (13) & 26.3040 (10) & 46.5434 (14) & 229.5613 (11) & 23.8719 (15) & 35.3885 (14) \\ \hline
12 & ss501        & 50.4477 (11) & 38.8727 (12) & 31.1372 (11) & 25.6216 (12) & 48.1135 (11) & 228.8920 (12) & 24.7030 (13) & 36.5725 (12) \\ \hline
13 & danielchoi   & 50.7946 (9)  & 39.1847 (10) & 31.4011 (10) & 25.8194 (11) & 48.1828 (10) & 226.1749 (13) & 24.8638 (11) & 36.8950 (11) \\ \hline
14 & Hi1988       & 48.9767 (13) & 37.1830 (14) & 29.2179 (14) & 23.4479 (14) & 47.8742 (13) & 217.7578 (14) & 24.0982 (14) & 36.4749 (13) \\ \hline
15 & BraveGirls   & 46.4913 (15) & 34.2065 (15) & 25.9219 (15) & 19.9941 (15) & 46.1757 (15) & 180.6635 (15) & 24.9094 (10) & 33.2203 (15) \\ \hline
16 & mobled37     & 39.8284 (16) & 27.7706 (16) & 20.0732 (16) & 14.7538 (16) & 39.5366 (16) & 134.2035 (16) & 18.9348 (16) & 27.7624 (16) \\ \hline
17 & rjsdn        & 35.2991 (17) & 23.4908 (17) & 15.9875 (17) & 10.9201 (17) & 37.0862 (17) & 113.8025 (17) & 16.9700 (17) & 25.3106 (17) \\ \hline
18 & snow0        & 31.7156 (20) & 19.2009 (19) & 12.2571 (18) & 7.9731 (18)  & 31.8328 (18) & 92.8288 (18)  & 14.5823 (19) & 22.6518 (18) \\ \hline
19 & soeun        & 27.3832 (25) & 16.1707 (23) & 10.1247 (22) & 6.4948 (21)  & 29.3049 (21) & 78.9487 (19)  & 12.9487 (22) & 20.3599 (21) \\ \hline
20 & do0ori       & 32.1687 (19) & 18.3737 (20) & 11.0992 (20) & 6.8483 (20)  & 30.1717 (20) & 77.8300 (20)  & 13.9382 (20) & 20.6390 (20) \\ \hline
21 & zhengmaozong & 33.7145 (18) & 19.5054 (18) & 11.7051 (19) & 7.0696 (19)  & 30.5224 (19) & 75.3119 (21)  & 15.0924 (18) & 20.8646 (19) \\ \hline
22 & yongsu23     & 26.9424 (27) & 15.3106 (25) & 9.3395 (24)  & 5.8594 (23)  & 28.2218 (22) & 73.1277 (22)  & 12.2745 (24) & 19.4324 (22) \\ \hline
23 & challang     & 30.8059 (21) & 17.0075 (22) & 9.8781 (23)  & 5.8459 (24)  & 27.7463 (23) & 65.9372 (23)  & 13.0716 (21) & 18.9231 (23) \\ \hline
24 & k1101jh      & 24.3603 (29) & 13.4100 (28) & 8.0403 (28)  & 5.0202 (27)  & 26.3008 (27) & 65.7418 (24)  & 11.2266 (28) & 17.6973 (24) \\ \hline
25 & gamy         & 28.6025 (23) & 15.3838 (24) & 9.0163 (25)  & 5.4843 (25)  & 26.9439 (25) & 64.0570 (25)  & 12.0974 (25) & 17.6462 (25) \\ \hline
26 & alsaco       & 27.0615 (26) & 14.4850 (27) & 8.4428 (26)  & 5.1177 (26)  & 26.6241 (26) & 64.0352 (26)  & 11.2880 (27) & 17.0905 (26) \\ \hline
27 & Yebin46      & 30.3189 (22) & 17.4146 (21) & 10.5292 (21) & 6.3576 (22)  & 27.6034 (24) & 59.7099 (27)  & 12.5039 (23) & 16.5475 (27) \\ \hline
28 & Rainy21      & 27.4143 (24) & 14.6171 (26) & 8.3537 (27)  & 4.9207 (28)  & 23.6640 (28) & 49.8930 (28)  & 11.7443 (26) & 15.9822 (28) \\ \hline
29 & Runbor       & 25.0107 (28) & 11.8485 (29) & 6.0223 (29)  & 3.1465 (29)  & 21.7858 (29) & 32.5173 (29)  & 9.5540 (29)  & 11.5799 (29) \\ \hline
30 & Merander     & 22.4721 (30) & 9.8276 (30)  & 4.7389 (30)  & 2.3512 (30)  & 20.1784 (30) & 26.2615 (30)  & 8.3881 (30)  & 10.4668 (30) \\ \hline
31 & kathmanducow & 15.9741 (31) & 6.4685 (31)  & 2.6371 (31)  & 1.1054 (31)  & 14.4777 (31) & 10.2777 (31)  & 5.7737 (31)  & 4.1072 (31)  \\ \hline
\end{tabular}
\end{adjustbox}
\caption{NICE 2023 Challenge results are presented. The numbers in the parentheses are individual ranking based on the metric of each column. The final rankings are determined by CIDEr score.}
\label{tab:results}
\end{table*}

\subsection{Dataset}\label{sec:chal_data}

The images and corresponding captions used in this challenge were provided by Shutterstock. This new large-scale evaluation dataset consists of approximately 26k high quality images, along with associated metadata. It includes a wide breadth of concepts from various categories. With this dataset, the participants of the challenge were expected to take a longitudinal evaluation across a variety of metrics to comparatively assess the performances of different zero-shot image captioning models. Some example images in NICE dataset are shown in Figure~\ref{fig:nice}. We did not provide specific set of training data for the challenge, as we aim for zero-shot image captioning, in which AI models can perform image captioning on the new data that was never seen in the training stage.

\subsection{Evaluation metrics}\label{sec:chal_eval}

There were several evaluation metrics used for this challenge. The evaluation metric with the first priority was CIDEr~\cite{vedantam2015cider} score, which is short for Consensus-based Image Description Evaluation. CIDEr score calculates the similarity score between two sentences by weighting each n-gram with the TF-IDF (Term Frequency Inverse Document Frequency) values. This scheme sets the weight of an n-gram higher if it appears in the sentence but not frequently in the whole set of captions, thus estimating the importance of that n-gram higher. CIDEr score is currently one of the most popular metrics for text comparison, and we chose it as the top priority evaluation metric for the challenge. In case of a tie, we used SPICE~\cite{anderson2016spice}, METEOR~\cite{banerjee2005meteor}, ROUGE~\cite{lin-2004-rouge}, Bleu~\cite{papineni2002bleu} in the order of priority.

\subsection{Challenge phases}\label{sec:chal_phase}

\textbf{Validation phase}: From February to April 2023, validation server was provided so that the participants could upload their prediction results and the server would calculate the scores by comparing them to the ground truth captions. In this phase, participants were able to access the validation data, which includes the image and the ground truth captions. As this is the first challenge on the task, we provided ground truth captions in order to let participants get acquainted with the data format and build strategies for the preparation of the final challenge entry.

\begin{figure*}[ht]
    \centering
    \includegraphics[width=\textwidth]{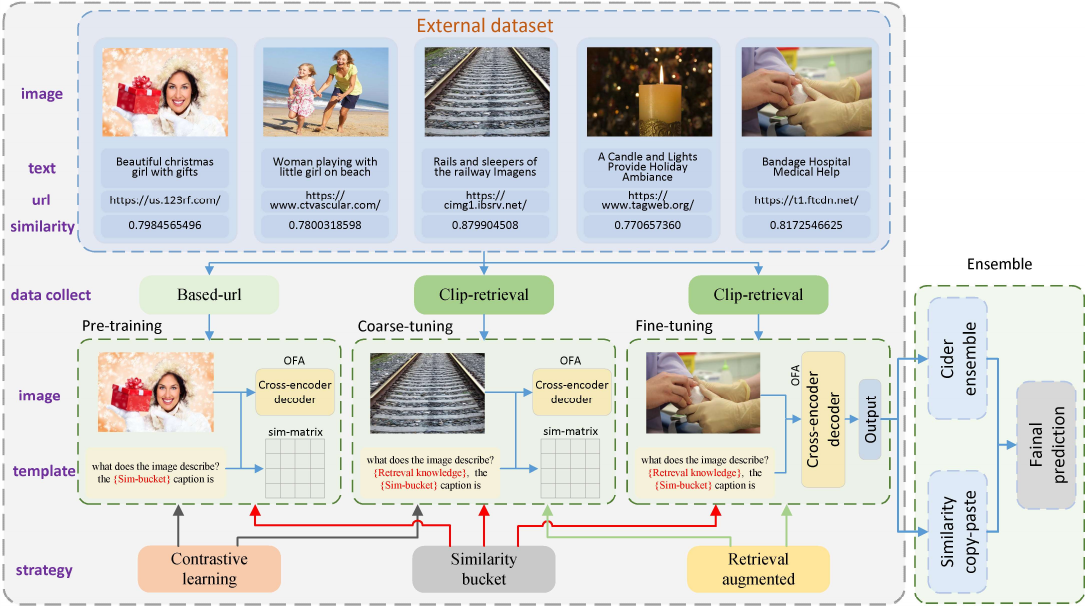}
    \caption{Overall Architecture of team "no". Their solution consists of four main stages, which includes Pre-training, Coarse-tuning, Fine-tuning and Model-ensemble. The training data for the first three stages are all collected from the large-scale Laion-5B dataset.}
    \label{fig:method_1}
\end{figure*}

\textbf{Test phase}: In April 2023, test server was open and allowed participants to upload prediction results up to 5 times during the phase. The score was calculated and the entry with the best CIDEr score was uploaded on the open leaderboard\footnote{\url{https://codalab.lisn.upsaclay.fr/competitions/10248\#results}}. In the test phase, the ground truth captions were not accessible and only the test scores were shown to the participants.

\subsection{Challenge results}\label{sec:chal_res}

The results of the challenge is presented in Table~\ref{tab:results}. There were 31 teams that participated in the challenge, and the final decision of the ranking was based on CIDEr score. The top-ranking entry scored 325.72, and the following entries scored 324.93, 316.23, and so on. Furthermore, the first ranking entry did not score the best in the other metrics, which shows that each entry had strong and weak points.


\section{Proposed Approaches}\label{sec:approaches}

\subsection{1st rank : no}\label{sec:approaches_1}

For the model level, they used OFA~\cite{wang2022ofa} as their base model. As shown in Figure~\ref{fig:method_1}, the overall architecture consists of three parts, namely Pre-training, Coarse-tuning and Fine-tuning. Pre-training stage aims to align a wide range of extensively visual concepts and store sufficient vision-language knowledge through contrastive learning, image captioning pre-train objectives. Coarse-tuning stage utilizes a small-scale external dataset similar to the competition domain, which can learn a large variety of novel concepts. Fine-tuning stage further compresses the dataset in the last stage and adds it to the competition validation dataset.

For the data level, they collected external training data from LAION-5B~\cite{schuhmann2022laion}, a large-scale CLIP-filtered image-text dataset. In the Pre-training stage, 1M image-text pairs are collected from the specific url(thumbx.shutterstock.com, editorial01.shutterstock.com, etc.). In the Coarse-tuning and Fine-tuning stage, they used all the competition images to retrieve external data from LAION-5B through Clip-retrieval\footnote{\url{https://github.com/rom1504/clip-retrieval}} library based on similarity. For each image query, they retrieved top-30 and top-10 image-text pairs, retain 120k samples for Coarse-tuning and 12k for Fine-tuning with the 5k validation dataset.

In addition, they introduced contrastive learning~\cite{li2021align}, similarity-bucket, retrieval-augmented~\cite{yang2023re} strategies to the model. Contrastive learning aims to learn better single-modal representations before fusion and align visual and text concepts in Pre-training stage. Similarity-bucket strategy  provides different similarity-prompts to the vision-language model in the all stages, which can control the model to generate the most matched and high-quality caption for a given image. Retrieval-augmented strategy provides a mini knowledge-base for each image-text pair during the training part. The model can not only extract visual features such as objects, attributes, and relationships of the image, but also explicitly align the information of the image with the knowledge in the knowledge-base.

\subsection{2nd rank : Retriever}\label{sec:approaches_2}

\begin{figure}[ht]
  \centering
   \includegraphics[width=\columnwidth]{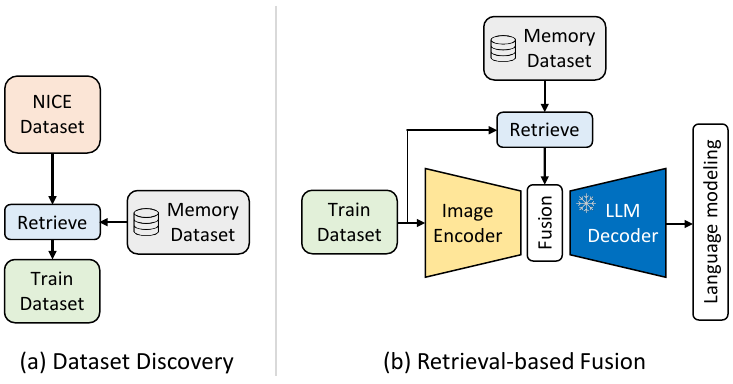}
   \caption{Overview of \textsc{Retriever} framework. It improves image captioning under zero-shot settings in two stages: \textbf{(a)} retrieval-based dataset discovery for training and \textbf{(b)} retrieval-based fusion conditioned on examples similar to input queries.
   }
   \label{fig:method_2}
\end{figure}

Captions in NICE dataset often contain new concepts like camera angle descriptions and proper nouns like place names, which are difficult to predict under zero-shot settings. Motivated from retrieval-augmented models~\cite{long2022retrieval,borgeaud2022improving}, the \textsc{Retriever} framework aims to complement such limited data condition by efficiently utilizing external knowledge in model training and inference. As shown in~\ref{fig:method_2}, it enhances typical captioning model in two stages. Firstly, to construct a dataset for training, an explicit retrieval module samples a set of image-text pairs from the memory, which can closely mimic the target distribution of captions. Secondly, the knowledge associated with the input query is explicitly combined into the model via retrieval-based fusion. These two processes are introduced as below, while the technical report~\cite{ohtechnical} includes more details.

\noindent \textbf{Dataset Discovery} 
The base captioning model BLIP-2~\cite{li2023blip} exhibits poor performance when evaluated on NICE dataset. This necessitates to discover data samples for training that are relevant to the target distribution. To that end, given a query image, a retrieval module is employed to retrieve a set of related examples from a dataset saved in the external memory via a $k$-nearest neighbor (kNN) search. This module is applied to the NICE dataset of size $N$, yielding total $kN$ images with corresponding captions from the external memory. After a deduplication step, the unique image-text pairs are utilized for further finetuning the BLIP-2.

\noindent \textbf{Retrieval-based fusion}
Given a query image either during training or inference, a set of value embeddings is produced by encoding the caption of the $k$ retrieved image-text pairs. These embeddings can provide rich contextual information complementary to the knowledge in the original model. They are aggregated and then concatenated with the query feature. After the fusion, the new feature is further passed to the remaining captioning pipeline; Q-Former followed by the LLM decoder in BLIP-2~\cite{li2023blip}. This results in more accurate captions, and the entire \textsc{Retriever} framework achieves CIDEr 324.9 on NICE test split. 

\subsection{3rd rank : Kakaobrain-MMU}\label{sec:approaches_3}

\begin{figure}[ht]
  \centering
   \includegraphics[width=\columnwidth]{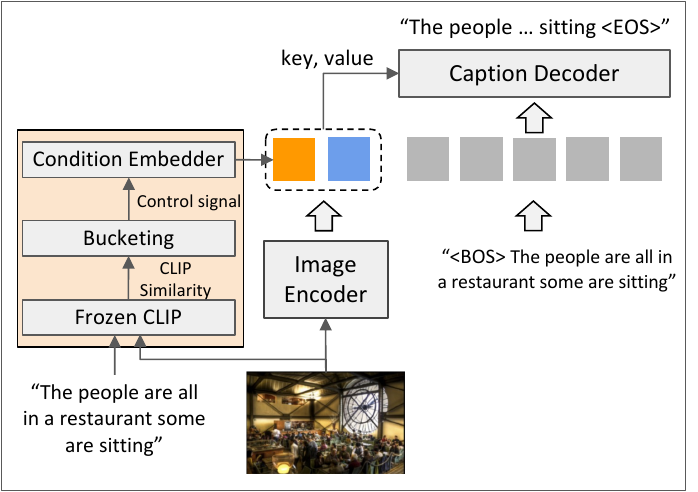}
   \caption{
        \small The architecture of captioning model. First, the control signal is computed by bucketing CLIP similarity during the training phase, and set to a constant one during the test phase. Then, it is concatenated to the image embeddings and is fed into a cross-attention-based caption decoder model as key-value features. This figure is inspired and slightly modified from \cite{noc}.
   }
   \label{fig:method_3}
\end{figure}

The main components of their approach are three-fold: 1) NoC framework~\cite{noc}, 2) three-stage training pipeline, and 3) consensus-based model ensemble.
Detailed explanations for each method will be described in the following sections.

\noindent \textbf{Training Algorithm}
Since they trained their models using large-scale web-crawled image-text paired datasets that contain inherent noises, \ie, misaligned pairs, they utilized their prior work, \textbf{\underbar{No}}ise-aware \textbf{\underbar{C}}aptioning (NoC) framework~\cite{noc}, as a primary training algorithm, illustrated in Figure~\ref{fig:method_3}. In a nutshell, NoC incorporates conditional modeling into a captioning model to model alignment levels of the input image-text pairs.
During the training phase, NoC utilizes the CLIP similarity between an image-text pair as an additional input of a control signal indicating the alignment level.
Then, at the inference phase, they feed a control signal of a desired alignment level into the captioning model to produce semantically accurate captions that are closely aligned with input images.

\begin{figure*}[ht]
  \centering
   \includegraphics[width=.9\textwidth]{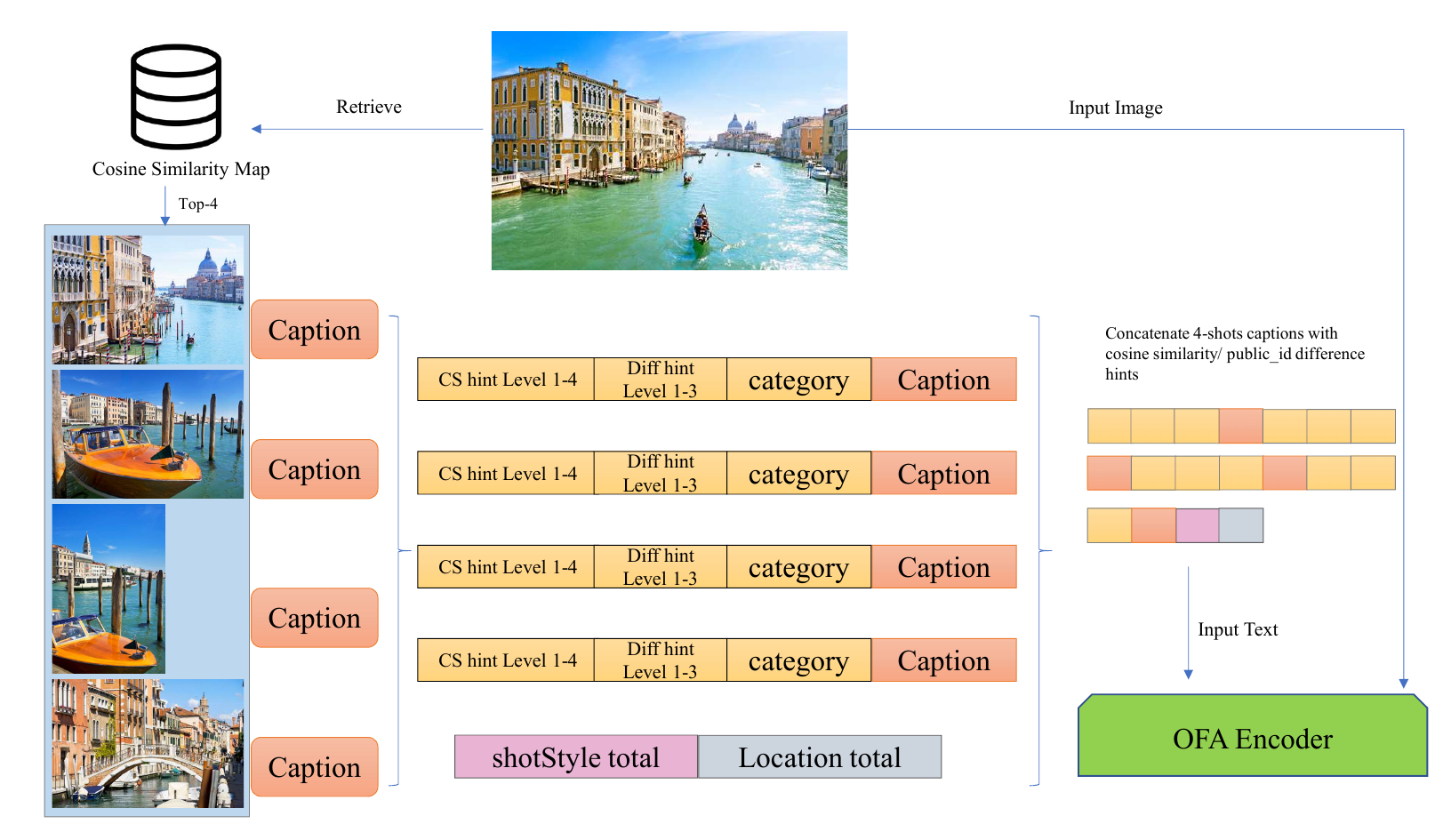}
   \caption{The overview of the method proposed by team Otsuka AI.}
   \label{fig:method_4}
\end{figure*}

\noindent \textbf{Three-stage Training}
Given web-crawled image-text data and NICE validation set, they trained their model with a three-stage pipeline: 1) pre-training on large-scale data, 2) fine-tuning on retrieved data, and 3) fine-tuning on NICE validation set. The model trained in each stage is used to initialize the model in the next stage. At the first stage, they pre-trained captioning models with the following datasets\footnote{They trained a captioning model for each dataset.}: CC15M~\cite{cc3m, cc12m}, COYO-700M~\cite{COYO}, COYO-100M (a subset of COYO-700M having CLIP similarity higher than 0.3), LAION-45M~\cite{schuhmann2022laion} (filtered from LAION-Aesthetics-en V1 52M), and 5) LAION-120M~\cite{schuhmann2022laion} (filtered from LAION-Aesthetics Predictor V2 600M). The filtering strategy for LAION-45M and -120M was based on aspect ratio, image size, CLIP similarity, text length, etc. Then they fine-tuned the pre-trained model using retrieved data. Inspired by \cite{react}, they retrieved the most NICE-relevant image-text pairs from the pre-training data. For each query sample (captions in NICE validation split), they retrieved 1,000 T2T and 1,000 T2I samples from the union of COYO and LAION datasets using CLIP ViT-B/32~\cite{CLIP} and FAISS~\cite{faiss}. This makes the captioning models likely to generate captions more similar to captions in NICE validation set. Finally, they further fine-tuned their captioning models from stage 2 with the NICE validation set to more tightly align the caption style with the ones in the NICE challenge. 

\noindent \textbf{Consensus-based Ranking for Ensemble}
Given $N$ captioning models, they generated a set of captions $\mathcal{C}_{N}=\{c_1,c_2,...,c_{N}\}$ for an input image where individual models generate a caption with a beam search strategy.
Then, for a caption $c_n$ from $n^{\text{th}}$ captioning model, they computed a consensus score~\cite{consensus1,consensus2} $s_n$ that is defined as an averaged similarity to all the other captions ($c_m \in {\mathcal{C}_{N}} \backslash \{ c_n \}$) as follows:
\begin{equation}
	s_n=\frac{1}{|\mathcal{C}_{N}|-1}\sum_{c_m \in {\mathcal{C}_{N}} \backslash \{ c_n \}} \text{sim}(c_n, c_m),\label{eq:ensemble}
\end{equation}
where $\text{sim}(c_n,c_m)$ is the CIDEr score between two captions $c_n$ and $c_m$.
Finally, the caption of the highest consensus score is chosen as their final output for the input image.

\subsection{4th rank : Otsuka AI}\label{sec:approaches_4}

In the NICE dataset, named entities such as geographical locations (e.g., Germany, Italy) pose a challenging task for inferring through images. For instance, deducing that the habitat of highland cattle is Germany based on pictures requires substantial foundational knowledge. In this work, problem is approached from the perspective of generating dialogue specialized for the NICE dataset, resembling persona conversations. The distinguished image captioning model OFA~\cite{wang2022ofa} is fine-tuned to be specialized for NICE.
 
The term "Levels" represents the entirety of the methodology, where soft prompts~\cite{qin2021learning} are effectively utilized to categorize the strength of hints present in example captions accessible by the model. Based on the output layer values of the OFA encoder, the captions of the four most similar images to the input image are retrieved from the pool of 5,000 validation images. Each caption serves as a powerful hint for generating predictions. To replicate the linguistic style of the NICE dataset, the four captions are employed akin to few-shot prompts. 

\begin{figure*}[ht]
  \centering
   \includegraphics[width=.9\textwidth]{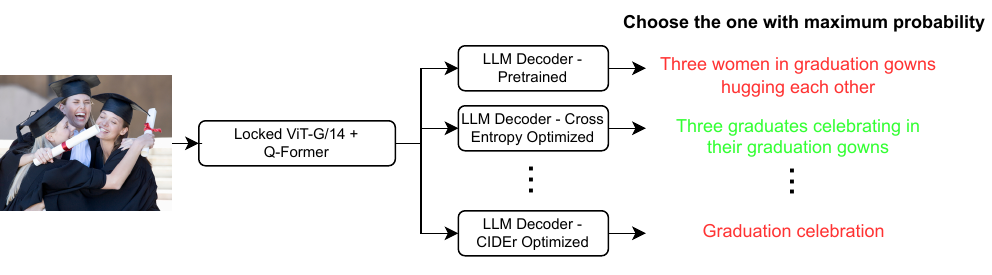}
   \caption{The proposed architecture from CLAS team.}
   \label{fig:method_5}
\end{figure*}

As a method to specify the level of hint in the prompt and facilitate the model's judgment, soft prompts are introduced. The values of cosine similarity between the input image and the four example images were normalized to provide their corresponding levels. These soft prompts are structured into four levels based on cosine similarity and three levels based on public id difference. Soft prompts form part of the broader utilization of various soft prompt techniques, as seen in visual prompting~\cite{jia2022visual}, symbol tuning~\cite{wei2023symbol}, and others.

\subsection{5th rank : CLAS}\label{sec:approaches_5}

Their approach revolves around the BLIP-2 architecture~\cite{li2023blip}, that combines several strategies to improve performance. The BLIP-2 architecture effectively utilizes image features by incorporating the Querying Transformer (Q-Former) along with state-of-the-art language models, such as OPT~\cite{zhang2022opt}. Q-Former is pretrained in two stages: the representation and generative learning stages, allowing the extraction of a fixed number of output features from the image encoder, regardless of the input image resolution. Their image encoder is based on ViT-G/14~\cite{fang2023eva}, while the language model is based on OPT-2.7b.

For fine-tuning, they used the validation set and the "a picture of" prompt, which helps the model converge faster and generate better captions. The network is fine-tuned on the validation set for 200 epochs. They employed FP16 mixed precision training and the Low-Rank Adaptation (LoRA) technique~\cite{hu2021lora} to allow for adaptation without updating all model parameters. After fine-tuning the network with cross-entropy loss, they utilized CIDEr optimization based on self-critical sequence training~\cite{rennie2017self}. This approach leverages the output of its own test-time inference algorithm to normalize the rewards it experiences.

Considering the high computational requirements of their model, conventional ensemble methods that average or rank sequences were not practical. Instead, they adopted an ensemble of 10 models, each trained with varying learning rates and epochs, using both cross-entropy and CIDEr optimization methods. To rank the models, they employed a model ranking system based on the confidence score of each generated caption. This confidence score is calculated by considering the probability of each word in the caption obtained through greedy sampling. The caption with the highest probability is selected for the output.

\subsection{6th rank : MKC}\label{sec:approaches_6}

\begin{figure}[ht]
\centering
    \includegraphics[width=.75\columnwidth]{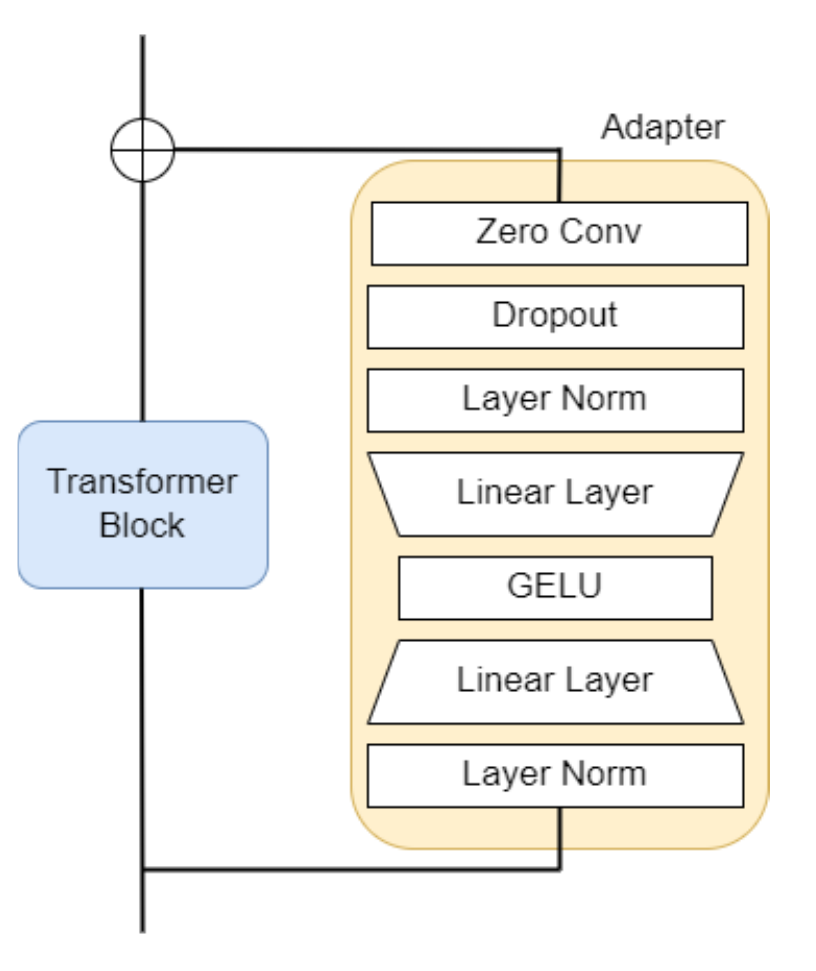}
  \caption{Proposed Adapter structure}
  \label{fig:method_6}
\end{figure}

They finetuned a large Vision-Language Pre-trained model, BLIP-2 ViT-g OPT 6.7B to cope with various concepts in the NICE dataset. To avoid catastrophic forgetting and reduce the number of parameters to train, they added adapters to BLIP-2 image encoder and only update the attached adapter, layernorm and Q-former using EMA(Exponential Moving Average) method. For finetuning, they divided the training process into two stages: 1) finetuning with the CC3M dataset and 2) finetuning with the NICE validation and the CC3M dataset mixed together.

\begin{figure*}[ht]
  \centering
   \includegraphics[width=\textwidth]{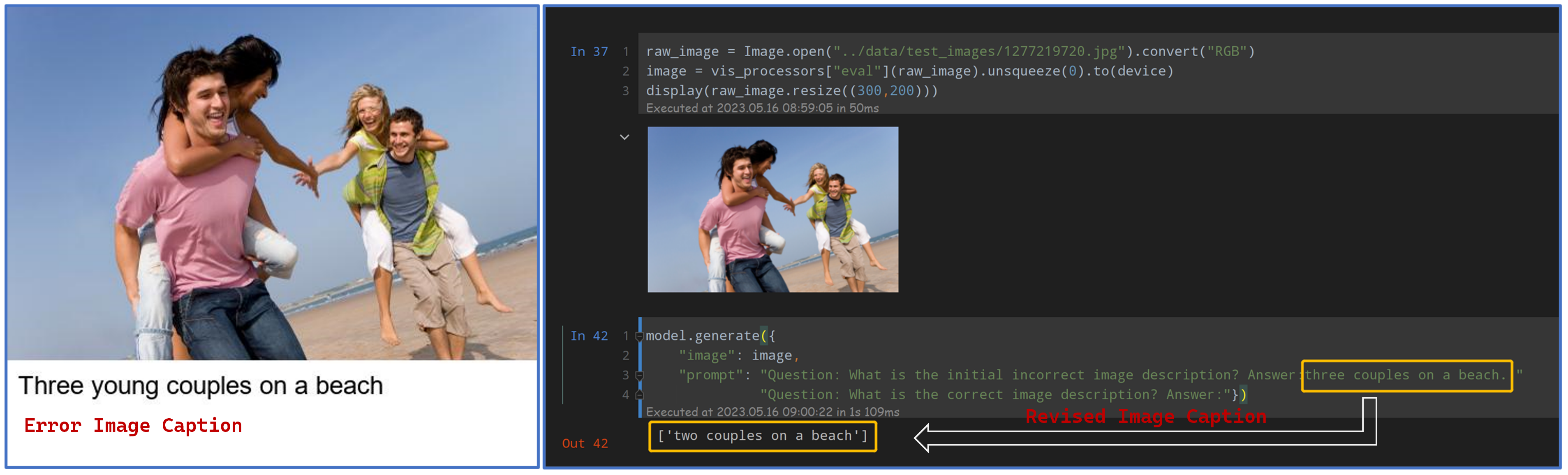}
   \caption{Instructed image caption generation.}
   \label{fig:method_7}
\end{figure*}

They plugged an adapter~\cite{kim2022large} with zero convolution~\cite{zhang2023adding} at each layer in ViT~\cite{fang2023eva}. Figure~\ref{fig:method_6} shows the structure of the adapter. The weights in zero convolutions would progressively grow from zero to optimal parameters, which helped the model slowly adapt to the CC3M dataset. They also optimized the layernorm~\cite{ba2016layer} inspired by the context of domain adaptation~\cite{song2023ecotta, niu2023towards}. Due to the various domains in NICE dataset, the method in domain adaptation could positively affect the performance. Lastly, Q-former in the original BLIP-2 was also optimized. The main role of the Q-former is to reduce the modality gap between image and natural language. Since the image feature extractor was decided to be trained, it was reasonable to optimize the Q-former as well. They also applied the EMA method to prevent catastrophic forgetting. They built a teacher model with parameter $\theta_{T}$, and regularized a student model, $\theta_{S}$, with consistency loss~\cite{tarvainen2017mean}. Consistency loss($Loss_{con}(\theta)$) between a teacher model($f(x;\theta_{T})$) and a student model($f(x;\theta_{S})$) was defined as follows:
\begin{align}
\mathcal{L}_{con}(\theta) = \beta  KL(f(x;\theta_{T}), f(x;\theta_{S}))
\end{align}
where $\beta$ is a hyperparameter for controlling the strength of consistency between a teacher and a student model, $x$ denotes an input image, and $f$ denotes the model architecture. The overall loss function for the whole finetuning process was defined as follows:
\begin{align}
\mathcal{L}_{total} = \mathcal{L}_{ce} + \mathcal{L}_{con}
\end{align}
where $\mathcal{L}_{ce}$ is a standard cross-entropy loss. At the first finetuning stage, only the CC3M dataset was used. Then in the second stage, the mixture of NICE validation dataset and the CC3M dataset was used.

In this work, they presented the method of light-weighted finetuning in the field of Vision-Language, especially aimed at the image captioning task. Choosing BLIP-2 as the baseline model, adapters with zero convolution and optimization of layer normalization were applied to perform efficient domain-robust finetuning. Their model achieved up to 274.69 CIDEr score on the test case, demonstrating the effectiveness of the proposed finetuning methodology.

\subsection{7th rank : Mtop}\label{sec:approaches_7}

Team Mtop's solution for the NICE 2023 challenge revolves around data augmentation and finetuning of the BEiT3~\cite{wang2022image} model with the augmented data. Due to the limitations in computational resources, they employed a minimal-effort approach. Firstly, they generated additional caption data by learning different styles. Next, they performed finetuning of the BEiT3 model using this augmented data without resorting to external datasets. To prevent catastrophic forgetting, a novel caption correction method was adopted. Their submission secured the 7th position in the challenge, achieving a notable score of 270 points.

\noindent \textbf{Details of Approach} BEiT3, pretrained on image-text pairs from datasets like CC12M, CC3M, SBU, COCO, and VG, served as the foundation for their approach. However, they recognized the significant differences in caption styles between these datasets and NICE, resulting in distributional discrepancies. To address this, they utilized BLIP2 to measure the similarity with NICE and selected image-text pairs that closely resembled the NICE dataset. These pairs enabled zero-shot learning during finetuning. One of the key challenges encountered during finetuning was catastrophic forgetting and being misled by the training data. To mitigate this issue, they employed instructed captions generated by large language models to guide the caption generation process. This approach led to improved performance and better results.

\noindent \textbf{Key Contributions} Their submission not only advances Image Captioning and Zero-shot Learning but also introduces a novel caption correction method to alleviate the impact of catastrophic forgetting. By finetuning the BEiT3 model with carefully selected augmented image-text pairs, they demonstrated the importance of considering target styles of captions for this task. Experimental results highlight the significance of caption correction and finetuning with relevant styles in achieving superior performance.


\section{Conclusion}\label{sec:conclusion}

Through NICE challenge 2023, a new dataset for zero-shot image captioning evaluation was proposed, and various approaches were attempted to appropriately adjust the AI models that had been trained on other datasets to the new evaluation dataset. We hope to continue this line of research and contribute to more challenging tasks to be performed by vision-language models. The proposed methods presented a wide range of insight on adapting the pretrained model to a specific domain of data, without sufficient training data from the target domain. This research field is expected to dive deeper into the real-world vision-language problems where the input visual data must be described in language of various styles.


{\small
\bibliographystyle{ieee_fullname}
\bibliography{egbib}
}


\appendix\label{sec:appendix}

\section{Organizers}
\noindent\textbf{Organizing committee}\newline
Seung Hwan Kim,\newline LG AI Research,\newline \texttt{sh.kim@lgresearch.ai}\newline
Kyoung Mu Lee,\newline Seoul National University,\newline \texttt{kyoungmu@snu.ac.kr}\newline
Bohyung Han,\newline Seoul National University,\newline \texttt{bhhan@snu.ac.kr}\newline
Alessandra Sala,\newline Shutterstock,\newline \texttt{asala@shutterstock.com}\newline

\noindent\textbf{Technical committee}\newline
Sihaeng Lee,\newline LG AI Research,\newline \texttt{sihaeng.lee@lgresearch.ai}\newline
Taehoon Kim,\newline LG AI Research,\newline \texttt{taehoon.kim@lgresearch.ai}\newline
Pyunghwan Ahn,\newline LG AI Research,\newline \texttt{p.ahn@lgresearch.ai}\newline
Sangyun Kim,\newline LG AI Research,\newline \texttt{syun.kim@lgresearch.ai}\newline
Mark Marsden,\newline Shutterstock,\newline \texttt{mmarsden@shutterstock.com}\newline

\section{Challenge participants}
\noindent\textbf{Team no}\newline
Xiangyu Wu,\newline Nanjing University of Science and Technology,\newline \texttt{wxy\_yyjhl@njust.edu.cn}\newline
Yi Gao,\newline Nanjing University of Science and Technology,\newline \texttt{gaoyi@njust.edu.cn}\newline
Hailiang Zhang,\newline Nanjing University of Science and Technology,\newline \texttt{121106022667@njust.edu.cn}\newline
YangYang,\newline Nanjing University of Science and Technology,\newline \texttt{yyang@njust.edu.cn}\newline
Weili Guo,\newline Nanjing University of Science and Technology,\newline \texttt{wlguo@njust.edu.cn}\newline
Jianfeng Lu,\newline Nanjing University of Science and Technology,\newline \texttt{lujf@njust.edu.cn}\newline

\noindent
\textbf{Team Retriever}\newline
Youngtaek Oh,\newline Korea Advanced Institute of Science and Technology,\newline \texttt{youngtaek.oh@kaist.ac.kr}\newline
Jae Won Cho,\newline Korea Advanced Institute of Science and Technology,\newline \texttt{chojw@kaist.ac.kr}\newline
Dong-Jin Kim,\newline Hanyang University,\newline \texttt{djdkim@hanyang.ac.kr}\newline
In So Kweon,\newline Korea Advanced Institute of Science and Technology,\newline \texttt{iskweon77@kaist.ac.kr}\newline
Junmo Kim,\newline Korea Advanced Institute of Science and Technology,\newline \texttt{junmo.kim@kaist.ac.kr}\newline

\noindent
\textbf{Team Kakaobrain-MMU}\newline
Wooyoung Kang,\newline Kakao Brain,\newline \texttt{edwin.kang@kakaobrain.com}\newline
Won Young Jhoo,\newline Kakao Brain,\newline \texttt{iji.young@kakaobrain.com}\newline
Byungseok Roh,\newline Kakao Brain,\newline \texttt{peter.roh@kakaobrain.com}\newline
Jonghwan Mun,\newline Kakao Brain,\newline \texttt{jason.mun@kakaobrain.com}\newline

\noindent
\textbf{Team Otsuka AI}\newline
Solgil Oh,\newline Wooribank,\newline \texttt{hhldtfrf@gmail.com}\newline

\noindent
\textbf{Team CLAS}\newline
Kenan Emir Ak,\newline Amazon Inc.,\newline \texttt{kenanea@amazon.com}\newline
Gwang-Gook Lee,\newline Amazon Inc.,\newline \texttt{gglee@amazon.com}\newline
Yan Xu,\newline Amazon Inc.,\newline \texttt{yanxuml@amazon.com}\newline
Mingwei Shen,\newline Amazon Inc.,\newline \texttt{mingweis@amazon.com}\newline

\noindent
\textbf{Team MKC}\newline
Kyomin Hwang,\newline Seoul National University,\newline \texttt{kyomin98@snu.ac.kr}\newline
Wonsik Shin,\newline Seoul National University,\newline \texttt{wonsikshin@snu.ac.kr}\newline
Kamin Lee,\newline Seoul National University,\newline \texttt{kamin.lee@snu.ac.kr}\newline
Wonhark Park,\newline Seoul National University,\newline \texttt{pwh0515@snu.ac.kr}\newline
Dongkwan Lee,\newline Seoul National University,\newline \texttt{biancco@snu.ac.kr}\newline
Nojun Kwak,\newline Seoul National University,\newline \texttt{nojunk@snu.ac.kr}\newline

\noindent
\textbf{Team Mtop}\newline
Yujin Wang,\newline Tsinghua University,\newline \texttt{yujeen.wang@gmail.com}\newline
Yimu Wang,\newline  University of Waterloo,\newline \texttt{yimu.wang@uwaterloo.ca}\newline
Tiancheng Gu,\newline University of Sydney,\newline \texttt{tigu8498@uni.sydney.edu.au}\newline
Xingchang Lv,\newline Illinois Institute of Technology,\newline \texttt{xlyu@hawk.iit.edu}\newline
Mingmao Sun,\newline University of California, Berkeley,\newline \texttt{sunmingmao@berkeley.edu}\newline

\end{document}